\documentclass[runningheads]{llncs}
\usepackage[T1]{fontenc}
\usepackage{graphicx}
\usepackage{listings}

\begin{document}
\title{Visual-Aware Representation of Web Pages for Machine Learning Applications}
%
%
\author{Radek Burget\inst{1}\orcidID{0000-0001-5233-0456} \and
Radek Hranický\inst{1}\orcidID{0000-0001-6315-8137}
}
\authorrunning{R. Burget and R. Hranický}
%
\institute{Brno University of Technology, Faculty of Information Technology, Bozetechova 2, 612\,00 Brno, Czechia\\
\email{\{burgetr,hranicky\}@fit.vut.cz}}
\maketitle              
\begin{abstract}
Applying machine learning to web pages is challenging due to the need to interpret HTML together with associated resources and perform rendering to obtain a meaningful visual and layout-aware representation. As a result, machine learning over web content remains comparatively underexplored. In this paper, we present a platform for visual-aware representation and machine learning over web pages based on the open-source rendering tool FitLayout. The platform provides a server capable of rendering web pages, explicitly capturing their visual and structural properties in an RDF-based representation, and persisting the rendered documents in an integrated storage. The processing pipeline is controlled via a REST API, while SPARQL queries are used to retrieve structured data suitable as input for machine learning algorithms. By explicitly modeling rendered web pages, including fine-grained layout details, the platform enables dataset sharing and supports the reproducibility of experimental results. The architecture supports the complete dataset preparation workflow, from web page collection and rendering through preprocessing and annotation of content elements to downstream learning tasks. We further provide a Python client library that integrates the platform with standard machine learning workflows. As a demonstration, we show how rendered web pages can be transformed into graph-based representations and used to train graph neural networks for recognizing key content elements, illustrating both the applicability of the approach and the reproducibility of the results.

\end{abstract}
\section{Introduction}
\label{sec:intro}
Machine learning for document understanding has progressed rapidly, particularly for visually rich inputs, such as scanned pages and PDFs. Many successful methods are layout-aware and represent documents as structured objects, often as graphs processed by graph neural networks (GNNs). Comparable approaches for web pages are still less explored.

Web pages are difficult because their layout is produced dynamically by a browser from HTML and external resources, and the rendering output has no standard, persistent representation. Consequently, most existing methods either analyze the DOM only (ignoring layout) or treat pages as images (losing content structure), limiting the use of layout-aware models.

Building on our previous work \cite{burget2023snapshots}, we use RDF descriptions of rendered pages as a machine-learning-ready representation and introduce a software platform based on our FitLayout open-source framework. The platform exposes an API and SPARQL access to the rendered representation, supports the sharing of rendered datasets, and provides a Python client for integration into common machine learning workflows.

We illustrate our approach using a case study on recognizing key content elements. We describe how to construct an annotated dataset of rendered web pages, and how to employ SPARQL queries to transform these pages into graph representations suitable for subsequent processing with GNNs. Finally, we explain how the resulting dataset can be shared to facilitate reproducibility of the results and to enable the application of alternative machine learning methods to the same underlying data.

\section{Related Work}
\label{sec:related}

Machine learning techniques have been effectively applied to identify key content elements in PDF documents, including table localization and structural parsing with neural methods designed for complex document layouts \cite{gemelli2022table,riba2019table} or complex layout analysis \cite{gemelli2023doc2graph,wang2023graphical}.

In contrast, web information extraction and web document understanding face additional challenges due to the dynamic, browser-rendered nature of web pages and the frequent mismatch between the DOM tree and the visual layout. Consequently, existing approaches often focus either on DOM or template-based extraction \cite{bevendorff2023dragnet,lin2020freedom,truong2026hybrid}, or incorporate visual cues by learning from screenshots/visual contexts \cite{gogar2016deep,kumar2022cova}, or from browser-derived render trees \cite{li2023wiert}.

This short paper extends our earlier work: in \cite{burget2023scraping}, we introduced an RDF-based model of rendered web pages to enable visually aware web scraping (in contrast to the more common DOM-based scrapers); in \cite{burget2023snapshots}, we presented a general method for creating snapshots of rendered web pages using this representation. In this study, we propose an entirely new approach that brings this solution into the Python-based machine learning ecosystem, enabling the use of frameworks such as PyTorch on web page data.

\section{Visual-Aware Representation of Web Pages}
\label{sec:representation}


To exploit the visual properties and layout of web page content in machine learning applications, the input web pages must be rendered, and the rendering results must be explicitly described. We implemented both tasks using the FitLayout framework \cite{burget2023snapshots}.

FitLayout allows the representation of a single rendered page at different levels of abstraction and provides an ontology suitable for this purpose. The ontology\footnote{\url{https://fitlayout.github.io/ontology/}} defines the concept of an {\em Artifact} that describes a single page on some level of abstraction. For machine learning tasks, the following two artifact types (subclasses of {\em Artifact}) are relevant:

\begin{itemize}
    \item The {\em Page} artifact describes a rendered page at the lowest level of abstraction as a set of {\em Boxes} (thus also called a box tree).
    \item The {\em AreaTree} represents an abstraction over the rendered page on the level of visible {\em visual areas} that are not directly connected to the source HTML code. 
\end{itemize}

In the {\em BoxTree}, every box is directly linked to a source HTML element, as specified by the CSS formatting model. Each box has visual attributes such as {\tt fontSize}, {\tt color}, {\tt positionX}, and {\tt positionY}. The root box corresponds to the entire page area, and hierarchical relationships are captured via the {\tt isChildOf} property. The {\em AreaTree} eliminates the rigid reliance on HTML markup by concentrating on visual areas, which can be identified using different techniques (such as page segmentation methods) provided by FitLayout. Finally, each area can be associated with one or more {\em Tags} via the {\em hasTag} property, which can be used to label areas during dataset preparation.

As a result, each rendered page is represented by two artifacts (a {\em Page} and an {\em AreaTree}), which correspond to RDF subgraphs stored in the FitLayout RDF repository.

\section{System Architecture and Processing Pipeline}
\label{sec:architecture}

The proposed system architecture is based on the observation that machine learning workloads are frequently spread across multiple computing nodes with distinct roles: storage nodes equipped with large-capacity storage for persisting data and compute nodes that provide sufficient processing power to carry out neural network training and evaluation.

\begin{figure}
\includegraphics[width=\textwidth]{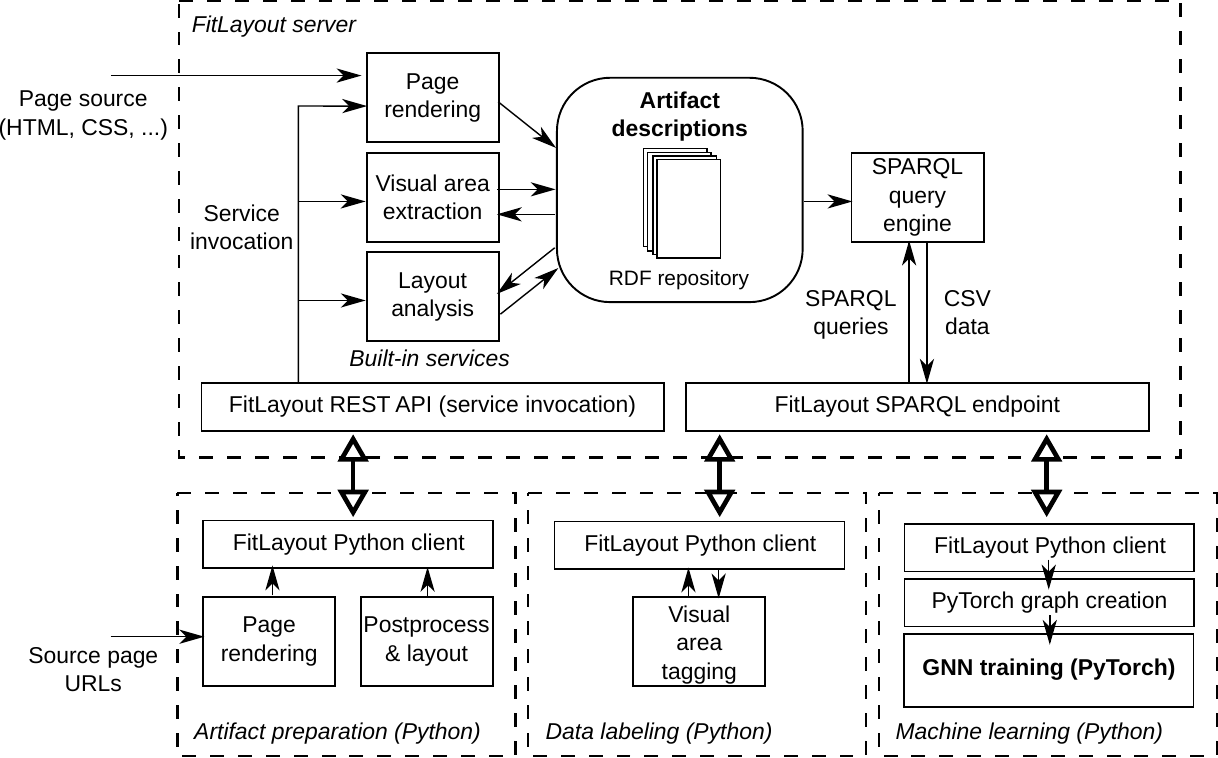}
\caption{The complete architectural layout, featuring the FitLayout server as the central data storage node and Python-based compute nodes responsible for dataset preparation as well as GNN training and evaluation.} \label{fig:arch}
\end{figure}

An overview of the entire architecture is shown in Fig. \ref{fig:arch}. It is composed of the FitLayout server, which acts as a data storage node with a central RDF repository for persisting page artifacts, and client-side Python applications that control dataset preparation and implement the machine learning methods.

To support the web page processing workflow, the server provides built-in services for creating different {\em artifacts}:

\begin{itemize}
    \item {\em Page rendering} -- it renders a web page in a headless Chromium web browser, which is controlled remotely via the Puppeteer library\footnote{\url{https://pptr.dev/}}. Then, the visual attributes of every rendered box are determined using JavaScript, and the {\em Page} RDF graph is created and stored in the repository.
    \item {\em Visual area extraction} -- for our task, we use a simple method that directly maps each visually distinct box in a {\em Page} to a visual area in the resulting {\em AreaTree} while discarding boxes that do not have any visual effect.
    \item {\em Layout analysis} -- examines the relative positions of sibling areas in the area tree and adds RDF statements indicating that one area is {\tt above}, {\tt below}, to the {\tt leftOf}, or to the {\tt rightOf} another area. FitLayout offers several implementations of this analysis; we employed an adapted version of the {\em visibility} method introduced by Gemelli et al. \cite{gemelli2022table}, originally designed for PDF documents.
\end{itemize}

Client applications can call these services through a REST API to generate and store RDF datasets of the rendered pages. Additionally, the server provides a SPARQL endpoint for querying the stored RDF data and inserting new statements. By default, FitLayout also includes an interactive client application with a web-based GUI\footnote{\url{https://github.com/FitLayout/PageView}}, which makes it possible to view and inspect the stored artifacts.

As Python has effectively become the standard environment for implementing machine learning methods, FitLayout offers a Python client library\footnote{\url{https://github.com/FitLayout/fitlayout-python-client}} for its REST API, enabling seamless integration of rendered page data into machine learning workflows.

\section{Python-Based ML Integration}
\label{sec:python}

The Python client layer, as shown in Fig. \ref{fig:arch}, covers two basic tasks: dataset preparation (which includes the preparation of the rendered page artifacts and their labeling) and the implementation of the machine learning applications themselves.

Artifact preparation involves sequentially calling the Page Rendering, Visual Area Extraction, and Layout Analysis services for each input page URL, with the goal of creating an {\em AreaTree} artifact for every source page. Each of these services is configurable; for instance, the client can set the viewport size used for rendering or use a specific method for spatial relationship discovery.

The optional data labeling step can be used to assign labels ({\em Tags}) to selected visual regions, which can later serve, for example, for training visual area classifiers. In a typical scenario, the target visual areas are first retrieved with a SELECT query, and subsequently, the {\em hasTag} statements are inserted into the RDF graph. The query can use both visual characteristics (e.g., color, size, or position) and properties of the underlying HTML, such as element attributes. For instance, in the Klarna web page dataset \cite{hotti2024klarna}, specific element attributes are employed to indicate the target elements; these attributes can then be straightforwardly converted into labels using this approach.

Finally, to implement the machine learning algorithms, the data on the visual areas used for training or testing can be efficiently retrieved from the repository via SPARQL SELECT queries. These queries make it possible to specify the target visual areas in the same way as in the previous step and select the relevant property values. For instance, when applying GNNs for visual area classification, the source graphs can be efficiently generated using one query to obtain node properties and another to obtain edge properties, as we illustrate in detail in the following section.


\section{Case Study: GNN-Based Content Element Recognition}
\label{sec:case}

To demonstrate the practical usefulness of the proposed architecture for machine learning on visual-aware web page models, we focus on a sample task: training a GNN-based classifier of visual areas to recognize book titles and prices in a fictional online bookstore\footnote{\url{https://books.toscrape.com/}}, a commonly used example target for web scraping. The entire project repository is available at GitHub\footnote{\url{https://github.com/FitLayout/graphlearn}}. The sample project covers the entire workflow from dataset preparation through the application of ML methods to exporting the dataset in a shareable format to support the reproducibility of the entire process.

\subsection{Environment setup}
\label{sec:case:setup}
The project contains client Python scripts that implement dataset preparation, data labeling, and machine learning, as described in the following subsections. The FitLayout server is set up by simply running the corresponding Docker image provided by the FitLayout project. For our experiments, we used a two-node configuration, where the server runs on a separate computing node with sufficient storage space. However, the client and server can share a single node when required.

\subsection{Dataset preparation}
The entire web page processing workflow starts by rendering the source pages based on a list of URLs. Collecting the input URLs is not covered here; we used a simple script to extract the URLs of pages corresponding to individual books in the bookstore. We collected 1,000 URLs in total and made them available in a text file within the project. The individual subtasks were implemented as short Python scripts that control the dataset preparation by simply calling the corresponding FitLayout services using the FitLayout Python client library:

{\em Page rendering} script ({\tt render.py}) invokes the Puppeteer-based rendering service for every input URL using the default viewport width of 1200 px. As a result, the corresponding {\em Page} artifacts are stored in the built-in RDF repository.

The {\em Postprocessing \& layout} script ({\tt segment.py}) applies the Visual area extraction service on each page, generating an {\em AreaTree} artifact. It then calls the layout analysis service to infer the spatial relationships between the visual areas, as detailed in Section \ref{sec:architecture}.

Finally, the {\em data labeling} script ({\tt tagging.py}) detects the visual areas corresponding to {\em book titles} and their {\em prices} in the repository and inserts the respective {\tt hasTag} statements into the RDF repository, which are subsequently used to train the GNN. In our basic use case, all source pages follow a shared template, enabling the identification of target regions through a single SPARQL query that combines visual features and source HTML element properties. For more heterogeneous input sources, multiple taggers would be required, or in the worst-case scenario, manual labeling would need to be carried out via the FitLayout GUI.

\subsection{GNN Training}

We implemented the GNN training using the widely adopted PyTorch Geometric (PyG) framework. For each AreaTree, we first build a corresponding PyG graph in which nodes represent visual areas, while edges capture both the parent-child relations in the tree and the spatial relations between sibling areas. Node features are retrieved via a SPARQL query and include text color, background color (RGB), X and Y pixel coordinates on the page, font size and weight, text length, and proportions of letters, digits, and punctuation in the contained text. The target class is encoded as a simple class index (1 for title, 2 for price, and 0 for all other areas). A separate SPARQL query is used to detect nodes that share either a parent-child or spatial relationship; these relationships are then encoded as edges in the PyG graph. In this manner, we create a PyTorch {\em dataset} that can serve as input for training the GNN.

As a representative GNN architecture, we employ the GCNConv operator from PyG and build a neural network with three convolutional layers to generate graph embeddings. The network consists of an input layer, a hidden layer with 128 channels, and an output layer with 10 channels, followed by a linear classifier. To train the network, we employed a straightforward training loop using the AdamW optimizer, cross-entropy as the loss function, and early stopping.

Our sample dataset, as outlined above, consists of 1,000 area trees containing a total of 37,768 visual areas. Although the primary aim of this sample project is to demonstrate how the proposed architecture can integrate rendered web pages into a Python-based ML workflow—and the underlying model is comparatively simple—the experiments resulted in only 0 to 4 misclassified visual areas across the entire set (depending on how the training and testing data were split). These results indicate that the proposed approach is indeed viable.

\subsection{Dataset sharing}

The FitLayout Python client also includes the capability to export and import the complete RDF repository in common RDF serialization formats, such as Turtle or N-Quads. The exported dataset captures all information about the rendered web pages and the generated artifacts (area trees), including their labels and even a screenshot of each page. This makes it possible to publish the dataset and/or reproduce the results even if the original web pages are altered or become unavailable. The dataset created as part of this case study is available at Zenodo\footnote{\url{https://zenodo.org/records/18674233}}.


\section{Conclusions}
\label{sec:conclusions}

The implemented platform demonstrates a comprehensive approach to integrating visual-aware representations of rendered web pages into machine learning workflows. By leveraging the FitLayout framework and RDF-based modeling, the system explicitly captures the visual and structural properties of web pages, enabling flexible data extraction and reproducible experiments. The provided Python client library facilitates seamless dataset preparation, labeling, and application of machine learning methods, such as graph neural networks for content element recognition, as illustrated in the case study. This architecture not only bridges the gap between web content and document understanding methods but also supports dataset sharing and reproducibility, addressing the key challenges in web page analysis using machine learning.

\subsubsection{Acknowledgements} This work was supported by the project Smart information technology for a resilient society, FIT-S-23-8209, funded by Brno University of Technology.

%
%
%
\bibliographystyle{splncs04}
\bibliography{refs}
\end{document}